\documentclass{article}
\usepackage{spconf,amsmath,graphicx}
\usepackage{amssymb,amsfonts}
\graphicspath{{figs/}}
\usepackage{booktabs}
\usepackage{etoolbox}
\usepackage{siunitx}
\usepackage{textcomp}
\usepackage{xcolor}
\usepackage{multirow}
\usepackage[caption=false, font=footnotesize]{subfig}
\usepackage[export]{adjustbox}

\usepackage{enumitem}
\usepackage{cite}

\title{ENCRYPTION INSPIRED ADVERSARIAL DEFENSE FOR VISUAL CLASSIFICATION}
\name{MaungMaung AprilPyone and Hitoshi Kiya}
\address{Tokyo Metropolitan University, Tokyo, Japan}
\begin{document}
%
\maketitle
\begin{abstract}
  Conventional adversarial defenses reduce classification accuracy whether or not a model is under attacks.
  Moreover, most of image processing based defenses are defeated due to the problem of obfuscated gradients.
  In this paper, we propose a new adversarial defense which is a defensive transform for both training and test images inspired by perceptual image encryption methods.
  The proposed method utilizes a block-wise pixel shuffling method with a secret key.
  The experiments are carried out on both adaptive and non-adaptive maximum-norm bounded white-box attacks while considering obfuscated gradients.
  The results show that the proposed defense achieves high accuracy (\SI{91.55}{\percent}) on clean images and (\SI{89.66}{\percent}) on adversarial examples with noise distance of $8/255$ on CIFAR-10 dataset.
  Thus, the proposed defense outperforms state-of-the-art adversarial defenses including latent adversarial training, adversarial training and thermometer encoding.
\end{abstract}
\section{Introduction}
\label{sec:intro}
Security in computer vision systems is quintessential and high in demand.
This is because computer vision technology has been deployed in many applications including safety and security critical applications such as self-driving cars, healthcare, facial recognition, etc.\ and many more visual recognition systems.
Computer vision systems are primarily powered by deep neural networks (DNNs).
It is proven that DNNs have brought impressive state-of-the-art results to computer vision.
However, researchers have already discovered that neural networks in general are vulnerable towards certain alteration in the input known as adversarial examples~\cite{Szegedy-ICLR-2014, Biggio-MLKDD-2013}.
These adversarial examples can cause neural networks misclassify or force to classify a targeted class with high confidence.
Incorrect decisions made by DNNs can cause serious and dangerous problems.
As an example, self-driving cars may misclassify ``Stop'' sign as ``Speed Limit''~\cite{Eykholt-CVPR-2018}.
Due to this threat, adversarial machine learning research has got a significant amount of attention recently although it has been started over a decade ago~\cite{Biggio-PR-2018}.

Researchers have proposed various attacks and defenses.
Ideally, provable robust models are desired.
Inspiring works such as~\cite{Raghunathan-ICLR-2018,Dvijotham-UAI-2018,Wong-ICML-2018} proposed provable secure training.
Although these methods are attractive and desirable, they are not available for larger datasets.
One recent work~\cite{Wong-NIPS-2018} scaled up to CIFAR-10~\cite{Krizhevsky-Report-2019} dataset in provable defense research.
However, the accuracy is not comparable even on low adversarial noise distance.
There is also an alternative approach to find a defensive transform $t(\cdot)$ so that the prediction of a classifier $f(\cdot)$ on clean image $x$ is equal to that of an adversarial example $x'$ (i.e., $f(x)=f(t(x'))$).
Such works include~\cite{Buckman-ICLR-2018,Samangouei-ICLR-2018,Guo-ICLR-2018,Xie-ICLR-2018,Song-ICLR-2018}, etc.
They all have been defeated when accounting for obfuscated gradients (a way of gradient masking)~\cite{Athalye-ICML-2018}.
To reinforce these weak defense methods, Raff et al.~\cite{Raff-CVPR-2019} proposed a stronger defense by combining a large number of transforms stochastically.
However, applying many transforms drop in accuracy even though the model is not under attack and is computationally expensive.
Our previous work removes adversarial noise generated on one-bit images by double quantization~\cite{Maung-Access-2019}, but, clean images are limited to be in one-bit.

Therefore, in this work, we propose a new adversarial defense which has been inspired by perceptual image encryption methods~\cite{Chuman-TIFS-2019,Warit-APSIPAT-2019,Tanaka-ICCETW-2018,Warit-Access-2019}.
It was reported that~\cite{Tanaka-ICCETW-2018} can be used as a defensive transform~\cite{Maung-GCCE-2019}.
However, it is not meant for adversarial defense and reduces accuracy.
To defend adversarial examples and maintain high accuracy, we design a defensive transform that uses a block-wise pixel shuffling method.
Similar to our work, Taran et al.\ proposed a key-based adversarial defense~\cite{Taran-ECCV-2018}.
The main intellectual differences include: (1) the proposed defense is inspired by perceptual image encryption (specifically, block-wise image encryption), in contrast to traditional cryptographic methods and (2) we consider white-box attacks unlike the work by~\cite{Taran-ECCV-2018} that considered gray-box attacks.
In an experiment, the proposed defense is confirmed to outperform state-of-the-art adversarial defenses including latent adversarial training, adversarial training and thermometer encoding under maximum-norm bounded threat model with the noise distance of $8/255$ on CIFAR-10 dataset.

\section{Preliminaries}
\subsection{Adversarial Examples}
An adversarial example is a modified input $x'$ (visually similar to $x$) to a classifier $f(\cdot)$ aiming $f(x) \neq f(x')$.
An attacker finds perturbation $\delta$ under certain distance metric (usually $\ell_p$ norm) to construct an adversarial example.
An attack algorithm usually minimizes the perturbation or maximizes the loss function, i.e.,
\begin{equation}
\underset{\delta}{\text{minimize}} \left\rVert \delta \right\rVert_p, \;\;\text{s.t.}\;\; f(x + \delta) \neq y, \text{or}
\end{equation}
\begin{equation}
  \underset{\delta \in \Delta}{\text{maximize}}\; \mathcal{L}(f(x + \delta), y),
\end{equation}
where $\Delta = \{\delta: \left\rVert \delta \right\rVert_p \leq \epsilon \}$.
There are many attack algorithms such as Fast Gradient Sign Method (FGSM)~\cite{Goodfellow-ICLR-2015}, Projected Gradient Descent (PGD)~\cite{Madry-ICLR-2018}, Carlini and Wagner (CW)~\cite{Carlini-SP-2017}, etc.

\subsection{Threat Model}
Following~\cite{Carlini-Arxiv-2019} and~\cite{Biggio-MLKDD-2013}, first, we describe a threat model that we use to evaluate the proposed defense.
We deploy PGD~\cite{Madry-ICLR-2018} because it is one of the strongest attacks under $\ell_\infty$ norm bounded metric.

Based on the goal of an adversary, the attack can be whether targeted ($f(x') = z$ where $z$ is a class targeted by the adversary) or untargeted ($f(x') \neq y$ where $y$ is a true class).
We focus on untargeted attacks under $\lVert x' -x \rVert_\infty \leq \epsilon$, where $\epsilon$ is a given noise distance.

We evaluate the proposed defense in white-box settings.
Therefore, we assume this adversary has full knowledge of the model, its parameters, trained weights, training data and the proposed defense mechanism except a secret key.

The adversary performs evasion attacks (i.e., test time attacks) in which small changes under $\ell_\infty$ metric change the true class of the input.
The adversary's capability is to modify the test image where the noise distance is $\epsilon$  in the range of $[2/255, 32/255]$.
Having full knowledge of the defense transform, our adversary also extends PGD.
Fully accounting obfuscated gradients, the adversary implements an adaptive attack like Backward Pass Differentiable Approximation (BPDA)~\cite{Athalye-ICML-2018} to estimate the correct gradients with a guessed key.


\section{Proposed Method}
\subsection{Overview}
\begin{figure}[htbp]
\centerline{\includegraphics{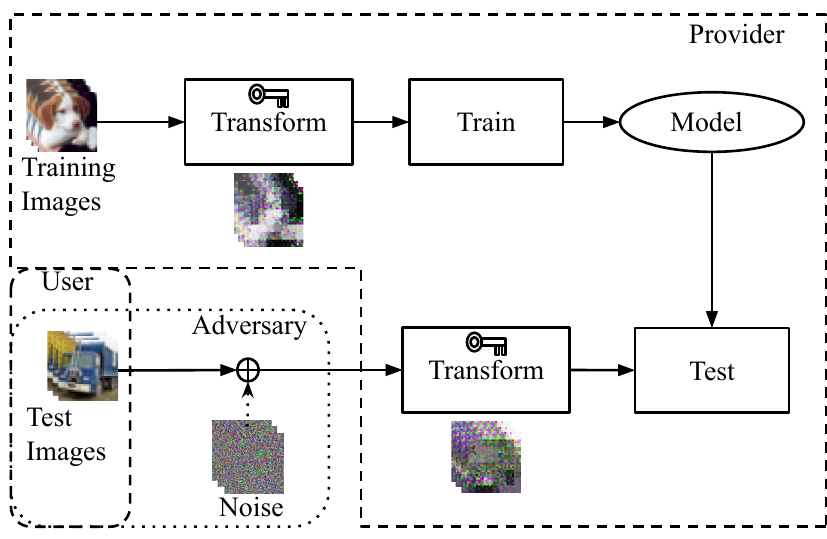}}
\caption{Overview of the proposed defense.}
\label{fig:icip2020-system}
\end{figure}

The goal of the proposed method is to hold high accuracy whether or not the model is under adversarial attacks.
The overview of the proposed defense is depicted in Fig.~\ref{fig:icip2020-system}.
Training images are transformed by a secret key and a model is trained by the transformed images.
Test images regardless of being clean images or adversarial examples are also transformed with the same key before classification process by the model.


\subsection{Defensive Transform}
We introduce a transform that exploits block-wise pixel shuffling with a secret key as an adversarial defense for the first time.
Both training and test images are transformed with a common key.
The transformation process is as follows.

A 3-channel (RGB), 8-bit image with a dimension of $X \times Y \times 3$ is divided into blocks (with the size of $M\times M \times 3$) where $X$ and $Y$ should be divisible by $M$.
Otherwise, padding is required.

Let $p(i)$ and $n$ be the pixel value and the number of pixels in each block (i.e., $M\times M \times 3$), where $i \in \{0, \ldots, n-1\}$.
The new pixel value $p'(i)$ is given by
  \begin{equation}
    p'(i) = p(\alpha(i)),
  \end{equation}
  where $\alpha = [\alpha(0), \alpha(1), \ldots, \alpha(n-2), \alpha(n-1)]$ is a random permutation vector of the integers from $0$ to $n-1$ generated by a key $K$.
\begin{figure}[htbp]
\centerline{\includegraphics{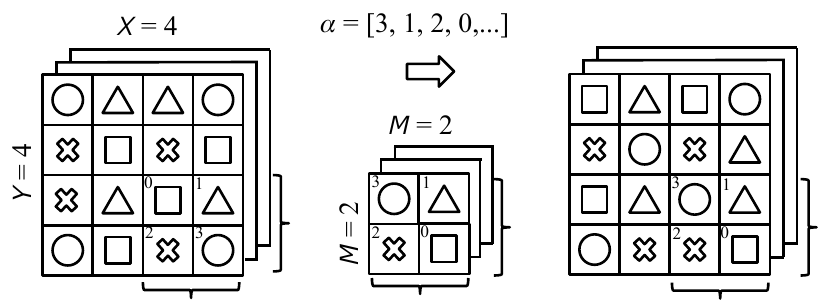}}
\caption{Process of block-wise pixel shuffling.}
\label{fig:key-transform}
\end{figure}
Fig.~\ref{fig:key-transform} illustrates the process of block-wise pixel shuffling.
The process is repeated for all the blocks in the image.

\subsection{Adaptive Attack}
\begin{figure}[htbp]
  \centerline{\includegraphics{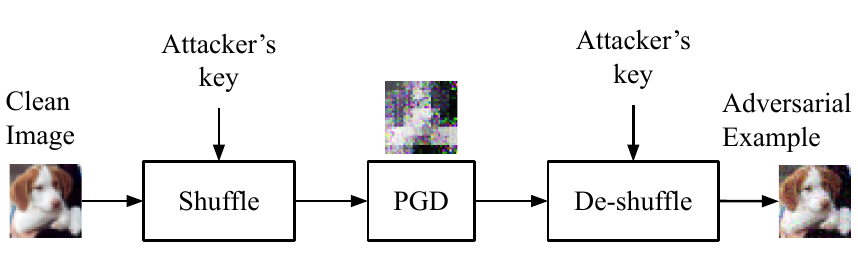}}
\caption{Diagram of adaptive attack.}
\label{fig:bpda}
\end{figure}

As pointed out by~\cite{Carlini-Arxiv-2019} and~\cite{Biggio-MLKDD-2013}, adaptive attacks are necessary in evaluating adversarial defenses.
Several recent defenses are defeated by adaptive attacks due to obfuscated gradients~\cite{Athalye-ICML-2018}.
To ensure the strength of the proposed defense, we implement a BPDA-like attack so that the gradients are correct with respect to the attacker's guessed key as shown in Fig.~\ref{fig:bpda}.
Basically, the adversary applies block-wise shuffling to a test image with a key, PGD is run on the shuffled image and the resulting adversarial example is de-shuffled with the adversary's assumed key.
We used random keys to attack the proposed method in our experiments.

\subsection{Key Management}
The proposed method uses a shared secret key $K$ to all the blocks in each of both training and test images.
Its key space is defined as follows:
\begin{equation}
  \mathcal{K}(n) = n!,
\end{equation}
where $n$ is the number of pixels in a block.
Deep learning is often done in the cloud server (provider) and the key $K$ should be saved securely at the server in deploying the proposed method.

\section{Experiments}
\subsection{Setup}
We used CIFAR-10~\cite{Krizhevsky-Report-2019} dataset with a batch size of 128 and live augmentation (random cropping with padding of 4 and random horizontal flip) on training set.
CIFAR-10 consists of 60,000 color images (dimension of $32 \times 32 \times 3$) with 10 classes (6000 images for each class) where 50,000 images are for training and 10,000 for testing.
Both training and test images were preprocessed by the proposed method with a common shared secret key $K$.

The deep residual network~\cite{He-CVPR-2016} with 18 layers (ResNet18) was trained for 160 epochs by the stochastic gradient descent optimizer.
The parameters are: momentum of $0.9$, weight decay of $0.0005$ and initial learning rate of $0.1$. 
A step learning rate scheduler was used with the settings (lr\_steps $= 40$, gamma $= 0.1$).

The parameters of PGD adversary are $\epsilon$ in the range of $[2/255, 32/255]$, and $\alpha = 2/255$.
The attack was run for $20$ and $40$ iterations with/without random initialization.
When random initializaition is set, perturbation is initialized with random values bounded by given $\epsilon$.

We used publicly available ResNet18 implementation~\cite{ResnetCode} on PyTorch.
The proposed method was implemented by modifying the code base of~\cite{Tanaka-ICCETW-2018}.
We deployed traditional PGD implementation from~\cite{AdverTorch-2019} and implemented BPDA-like attack to make the adversary adaptive and effective.

\subsection{Results}
\subsubsection{PGD Attack on Various Block Sizes}
We evaluated the proposed method under the use of various block sizes, $M \in \{2, 4, 8, 16\}$ by PGD.
We trained ResNet18 with images transformed by the proposed method with different block size $M \in \{2, 4, 8, 16\}$ resulting four models.
The trained models were first attacked by PGD with $\epsilon = 32/255$ for $20$ iterations (i.e., PGD$_{20}$) without random initialization.

Table~\ref{tab:block-size} summarizes the results obtained from the experiment of the proposed method.
The model trained with transformed images where $M=2$ gave the best performance (\SI{94.08}{\percent}) when the model is not under attacks.
However, $M=4$ performed better under attacks (i.e., \SI{84.72}{\percent}).
The results suggest that $M=4$ provides the best overall performance.


\robustify\bfseries
\sisetup{table-parse-only,detect-weight=true,detect-inline-weight=text,round-mode=places,round-precision=4}
\begin{table}[tbp]
  \caption{Accuracy of the proposed method under the use of various block sizes on PGD$_{20}$ ($\epsilon = 32/255$)\label{tab:block-size}}
  \centering
  \begin{tabular}{lSS}
  \toprule
  {$M \times M$} & {Clean} & {PGD$_{20}$}\\
  \midrule
  $2 \times 2$ & \bfseries \num{0.9408} & 0.7157\\
  $4 \times 4$ & 0.9155 & \bfseries \num{0.8472}\\
  $8 \times 8$ & 0.854 & 0.7892\\
  $16 \times 16$ & 0.7351 & 0.6756\\
  \bottomrule
  \end{tabular}
\end{table}


\sisetup{table-parse-only,detect-weight=true,detect-inline-weight=text,round-mode=places,round-precision=2}
\subsubsection{PGD Attack in Various Settings}
We further ran PGD attacks with various settings to the model trained by the proposed defense where $M = 4$.
The attacks were executed for $20$ and $40$ iterations, and subscript $r$ denotes random initialization (e.g., PGD$_{20r}$ stands for PGD attack for $20$ iterations with random initialization and BPDA denotes the adaptive attack).

Table~\ref{tab:block4-untargeted} captures the results of untargeted attacks where $\epsilon = 8/255$ and $32/255$.
When $\epsilon = 8/255$, the model maintain \SI{91.55}{\percent} accuracy on clean images and \SI{89.66}{\percent} on BPDA$_{40r}$ attack.
This confirms that the adaptive attack cannot reduce the accuracy when the attacker's key is not correct.
However, when $\epsilon$ was increased to $32/255$, BPDA$_{40r}$ reduced the accuracy to \SI{61.60}{\percent}.

\begin{figure}[htbp]
  \centerline{\includegraphics[width=\linewidth]{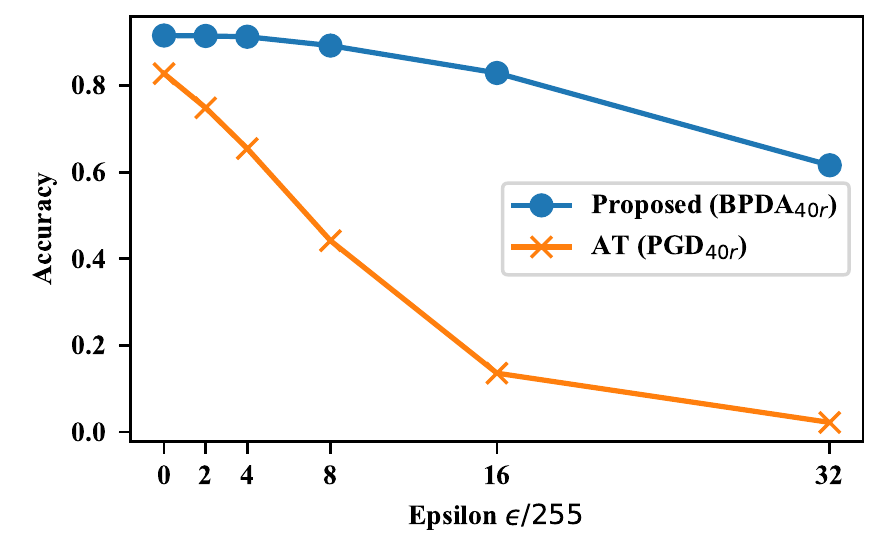}}
  \caption{Accuracy vs. perturbation budget.}
\label{fig:acc-epsilon}
\end{figure}

Our experiments show that BPDA$_{40r}$ is a better adversary.
Therefore, we evaluated the proposed defense with various $\epsilon \in \{2/255, 4/255, 8/255, 16/255, 32/255\}$ by BPDA$_{40r}$.
Moreover, to confirm the effectiveness of the proposed method, we also implemented state-of-the-art adversarial defense method, i.e., adversarial training (AT)~\cite{Madry-ICLR-2018} on the same network specifications with $\epsilon = 8/255$ to compare the results.
The accuracy versus various noise distances is plotted in Fig.~\ref{fig:acc-epsilon}.
When $\epsilon < 8/255$, the model trained by the proposed defense provides more than \SI{90}{\percent} accuracy.
The accuracy gradually drops when $\epsilon$ is greater than $8/255$.
Specifically, when $\epsilon = 16/255$, the model achieves $\approx$ \SI{83}{\percent} accuracy.
On the worst case scenario (i.e. $\epsilon = 32/255$), the accuracy of the model is $\approx$ \SI{62}{\percent}.
Nevertheless, the proposed method outperforms AT in any given perturbation budget as shown in Fig.~\ref{fig:acc-epsilon}.

\sisetup{table-parse-only,detect-weight=true,detect-inline-weight=text,round-mode=places,round-precision=4}
\begin{table*}[tbp]
  \caption{Accuracy of the proposed method ($M=4$) by PGD attack in various settings\label{tab:block4-untargeted}}
  \centering
  \begin{tabular}{lSSSSSSSSS}
  \toprule
  {Epsilon $\epsilon$} & {Clean} & {PGD$_{20}$} & {PGD$_{20r}$} & {PGD$_{40}$} & {PGD$_{40r}$} & {BPDA$_{20}$} & {BPDA$_{20r}$} & {BPDA$_{40}$} & {BPDA$_{40r}$}\\
  \midrule
  $8/255$ & \bfseries \num{0.9155} & 0.8948 & 0.8977 & 0.8917 & 0.8931 & 0.8988 & 0.8988 & 0.8988 & \bfseries \num{0.8966}\\
  $32/255$ & \bfseries \num{0.9155} & 0.8472 & 0.6645 & 0.777 & 0.6323 & 0.8204 & 0.6505 & 0.7346 & 0.616\\
  \bottomrule
  \end{tabular}
\end{table*}



\sisetup{table-parse-only,detect-weight=true,detect-inline-weight=text,round-mode=places,round-precision=2}

\subsection{Comparison with State-of-the-art Defenses}
To confirm the effectiveness of the proposed defense, we made a comparison with state-of-the-art published defenses for CIFAR-10 dataset on RobustML catalog\footnote{https://www.robust-ml.org/}.
We compared the proposed defense with the recent three defenses: latent adversarial training (LAT)~\cite{Kumari-IJCAI-2019}, adversarial training (AT)~\cite{Madry-ICLR-2018} and thermometer encoding (TE)~\cite{Buckman-ICLR-2018}.
All three defenses used wide residual network~\cite{Sergey-BMVC-2016} and were evaluated on $\ell_{\infty}$ threat model with $\epsilon = 8/255$ except LAT (used $\epsilon = 0.03$).
Table~\ref{tab:comparison} shows the summary of the comparison.
The proposed model was trained on ResNet18 and achieves superior accuracy (i.e., \SI{91.55}{\percent} on clean images and \SI{89.66}{\percent} on attacked ones).
Even on the worst case scenario (i.e., $\epsilon = 32/255$), the accuracy of the proposed method was still higher than the state-of-the-art defenses whether or not the model was under attacks.

\begin{table}[tbp]
\caption{Comparison with state-of-the-art defenses on CIFAR-10 dataset\label{tab:comparison}}
  \centering
  \resizebox{\columnwidth}{!}{%
  \begin{tabular}{llSS}
  \toprule
  {Defense} & {Threat Model} & {Clean} & {Attacked}\\
  \midrule
  LAT~\cite{Kumari-IJCAI-2019} & $\ell_\infty (\epsilon = 0.03)$ & 87.8  & 53.82\\
  AT~\cite{Madry-ICLR-2018} & $\ell_\infty (\epsilon = 8/255)$ & 87. & 46.\\
  TE~\cite{Buckman-ICLR-2018} & $\ell_\infty (\epsilon = 8/255)$ & 90. & 30.\\
  Proposed & $\ell_\infty (\epsilon = 8/255)$ & \bfseries \num{91.55} & \bfseries \num{89.66}\\
  Proposed & $\ell_\infty (\epsilon = 16/255)$ & \bfseries \num{91.55} & \bfseries \num{82.9}\\
Proposed & $\ell_\infty (\epsilon = 32/255)$ & \bfseries \num{91.55} & \bfseries \num{61.6}\\
  \bottomrule
  \end{tabular}
}
\end{table}



\section{Conclusion}
In this paper, we proposed a new adversarial defense that utilizes a key-based block-wise pixel shuffling method as a defensive transform for the first time.
Specifically, both training and test images are transformed by the proposed method with a common key before training and testing.
We also implemented an adaptive attack to verify the strength of the proposed defense.
Our experiments suggest that the proposed defense is resistant to both adaptive and non-adaptive attacks.
The results show that the proposed defense achieves higher accuracy, \SI{91.55}{\percent} on clean images and \SI{89.66}{\percent} on adversarial examples.
Compared to state-of-the-art defenses, the accuracy of the proposed method is \SI{35.84}{\percent} better than latent adversarial training, \SI{43.66}{\percent} than adversarial training and \SI{59.66}{\percent} than thermometer encoding under a maximum-norm bounded white-box threat model with the noise distance of $8/255$ on CIFAR-10 dataset.


\begin{small}
\bibliographystyle{IEEEbib}
\bibliography{ref}

\begin{thebibliography}{10}

\bibitem{Szegedy-ICLR-2014}
Christian Szegedy, Wojciech Zaremba, Ilya Sutskever, Joan Bruna, Dumitru Erhan,
  Ian~J. Goodfellow, and Rob Fergus,
\newblock ``Intriguing properties of neural networks,''
\newblock in {\em International Conference on Learning Representations}, 2014.

\bibitem{Biggio-MLKDD-2013}
Battista Biggio, Igino Corona, Davide Maiorca, Blaine Nelson, Nedim
  {\v{S}}rndi{\'c}, Pavel Laskov, Giorgio Giacinto, and Fabio Roli,
\newblock ``Evasion attacks against machine learning at test time,''
\newblock in {\em Joint European conference on machine learning and knowledge
  discovery in databases}. Springer, 2013, pp. 387--402.

\bibitem{Eykholt-CVPR-2018}
Kevin Eykholt, Ivan Evtimov, Earlence Fernandes, Bo~Li, Amir Rahmati, Chaowei
  Xiao, Atul Prakash, Tadayoshi Kohno, and Dawn Song,
\newblock ``Robust physical-world attacks on deep learning visual
  classification,''
\newblock in {\em Proceedings of the IEEE Conference on Computer Vision and
  Pattern Recognition}, 2018, pp. 1625--1634.

\bibitem{Biggio-PR-2018}
Battista Biggio and Fabio Roli,
\newblock ``Wild patterns: Ten years after the rise of adversarial machine
  learning,''
\newblock {\em Pattern Recognition}, vol. 84, pp. 317--331, 2018.

\bibitem{Raghunathan-ICLR-2018}
Aditi Raghunathan, Jacob Steinhardt, and Percy Liang,
\newblock ``Certified defenses against adversarial examples,''
\newblock in {\em International Conference on Learning Representations}, 2018.

\bibitem{Dvijotham-UAI-2018}
Krishnamurthy Dvijotham, Robert Stanforth, Sven Gowal, Timothy~A Mann, and
  Pushmeet Kohli,
\newblock ``A dual approach to scalable verification of deep networks.,''
\newblock in {\em UAI}, 2018, vol.~1, p.~2.

\bibitem{Wong-ICML-2018}
Eric Wong and J.~Zico Kolter,
\newblock ``Provable defenses against adversarial examples via the convex outer
  adversarial polytope,''
\newblock in {\em Proceedings of the 35th International Conference on Machine
  Learning}, 2018, pp. 5283--5292.

\bibitem{Wong-NIPS-2018}
Eric Wong, Frank Schmidt, Jan~Hendrik Metzen, and J~Zico Kolter,
\newblock ``Scaling provable adversarial defenses,''
\newblock in {\em Advances in Neural Information Processing Systems}, 2018, pp.
  8400--8409.

\bibitem{Krizhevsky-Report-2019}
Alex Krizhevsky,
\newblock ``Learning multiple layers of features from tiny images,''
\newblock Tech. {R}ep., 2009.

\bibitem{Buckman-ICLR-2018}
Jacob Buckman, Aurko Roy, Colin Raffel, and Ian Goodfellow,
\newblock ``Thermometer encoding: One hot way to resist adversarial examples,''
\newblock in {\em International Conference on Learning Representations}, 2018.

\bibitem{Samangouei-ICLR-2018}
Pouya Samangouei, Maya Kabkab, and Rama Chellappa,
\newblock ``Defense-{GAN}: Protecting classifiers against adversarial attacks
  using generative models,''
\newblock in {\em International Conference on Learning Representations}, 2018.

\bibitem{Guo-ICLR-2018}
Chuan Guo, Mayank Rana, Moustapha Cisse, and Laurens van~der Maaten,
\newblock ``Countering adversarial images using input transformations,''
\newblock in {\em International Conference on Learning Representations}, 2018.

\bibitem{Xie-ICLR-2018}
Cihang Xie, Jianyu Wang, Zhishuai Zhang, Zhou Ren, and Alan Yuille,
\newblock ``Mitigating adversarial effects through randomization,''
\newblock in {\em International Conference on Learning Representations}, 2018.

\bibitem{Song-ICLR-2018}
Yang Song, Taesup Kim, Sebastian Nowozin, Stefano Ermon, and Nate Kushman,
\newblock ``Pixeldefend: Leveraging generative models to understand and defend
  against adversarial examples,''
\newblock in {\em International Conference on Learning Representations}, 2018.

\bibitem{Athalye-ICML-2018}
Anish Athalye, Nicholas Carlini, and David~A. Wagner,
\newblock ``Obfuscated gradients give a false sense of security: Circumventing
  defenses to adversarial examples,''
\newblock in {\em Proceedings of the 35th International Conference on Machine
  Learning}, 2018, pp. 274--283.

\bibitem{Raff-CVPR-2019}
Edward Raff, Jared Sylvester, Steven Forsyth, and Mark McLean,
\newblock ``Barrage of random transforms for adversarially robust defense,''
\newblock in {\em Proceedings of the IEEE Conference on Computer Vision and
  Pattern Recognition}, 2019, pp. 6528--6537.

\bibitem{Maung-Access-2019}
MaungMaung AprilPyone, Yuma Kinoshita, and Hitoshi Kiya,
\newblock ``Adversarial robustness by one bit double quantization for visual
  classification,''
\newblock {\em IEEE Access}, vol. 7, pp. 177932--177943, 2019.

\bibitem{Chuman-TIFS-2019}
Tatsuya Chuman, Warit Sirichotedumrong, and Hitoshi Kiya,
\newblock ``Encryption-then-compression systems using grayscale-based image
  encryption for jpeg images,''
\newblock {\em IEEE Transactions on Information Forensics and Security}, vol.
  14, no. 6, pp. 1515--1525, 2019.

\bibitem{Warit-APSIPAT-2019}
Warit Sirichotedumrong and Hitoshi Kiya,
\newblock ``Grayscale-based block scrambling image encryption using ycbcr color
  space for encryption-then-compression systems,''
\newblock {\em APSIPA Transactions on Signal and Information Processing}, vol.
  8, 2019.

\bibitem{Tanaka-ICCETW-2018}
Masayuki Tanaka,
\newblock ``Learnable image encryption,''
\newblock in {\em 2018 IEEE International Conference on Consumer
  Electronics-Taiwan (ICCE-TW)}. IEEE, 2018, pp. 1--2.

\bibitem{Warit-Access-2019}
Warit Sirichotedumrong, Yuma Kinoshita, and Hitoshi Kiya,
\newblock ``Pixel-based image encryption without key management for
  privacy-preserving deep neural networks,''
\newblock {\em IEEE Access}, vol. 7, pp. 177844--177855, 2019.

\bibitem{Maung-GCCE-2019}
MaungMaung AprilPyone, Warit Sirichotedumrong, and Hitoshi Kiya,
\newblock ``Adversarial test on learnable image encryption,''
\newblock in {\em 2019 IEEE 8th Global Conference on Consumer Electronics
  (GCCE)}, 2019, pp. 693--695.

\bibitem{Taran-ECCV-2018}
Olga Taran, Shideh Rezaeifar, and Slava Voloshynovskiy,
\newblock ``Bridging machine learning and cryptography in defence against
  adversarial attacks,''
\newblock in {\em Proceedings of the European Conference on Computer Vision
  (ECCV)}, 2018.

\bibitem{Goodfellow-ICLR-2015}
Ian~J. Goodfellow, Jonathon Shlens, and Christian Szegedy,
\newblock ``Explaining and harnessing adversarial examples,''
\newblock in {\em International Conference on Learning Representations}, 2015.

\bibitem{Madry-ICLR-2018}
Aleksander Madry, Aleksandar Makelov, Ludwig Schmidt, Dimitris Tsipras, and
  Adrian Vladu,
\newblock ``Towards deep learning models resistant to adversarial attacks,''
\newblock in {\em International Conference on Learning Representations}, 2018.

\bibitem{Carlini-SP-2017}
Nicholas Carlini and David Wagner,
\newblock ``Towards evaluating the robustness of neural networks,''
\newblock in {\em 2017 IEEE Symposium on Security and Privacy (SP)}. IEEE, May
  2017, pp. 39--57.

\bibitem{Carlini-Arxiv-2019}
Nicholas Carlini, Anish Athalye, Nicolas Papernot, Wieland Brendel, Jonas
  Rauber, Dimitris Tsipras, Ian Goodfellow, and Aleksander Madry,
\newblock ``On evaluating adversarial robustness,''
\newblock {\em arXiv preprint arXiv:1902.06705}, 2019.

\bibitem{He-CVPR-2016}
Kaiming He, Xiangyu Zhang, Shaoqing Ren, and Jian Sun,
\newblock ``Deep residual learning for image recognition,''
\newblock in {\em Proceedings of the IEEE conference on computer vision and
  pattern recognition}, 2016, pp. 770--778.

\bibitem{ResnetCode}
Kuangliu,
\newblock ``Train cifar10 with pytorch,''
  https://github.com/kuangliu/pytorch-cifar, 2017.

\bibitem{AdverTorch-2019}
Gavin~Weiguang Ding, Luyu Wang, and Xiaomeng Jin,
\newblock ``{AdverTorch} v0.1: An adversarial robustness toolbox based on
  pytorch,''
\newblock {\em arXiv preprint arXiv:1902.07623}, 2019.

\bibitem{Kumari-IJCAI-2019}
Nupur Kumari, Mayank Singh, Abhishek Sinha, Harshitha Machiraju, Balaji
  Krishnamurthy, and Vineeth~N Balasubramanian,
\newblock ``Harnessing the vulnerability of latent layers in adversarially
  trained models,''
\newblock in {\em Proceedings of the 28th International Joint Conference on
  Artificial Intelligence}, 7 2019, pp. 2779--2785.

\bibitem{Sergey-BMVC-2016}
Sergey Zagoruyko and Nikos Komodakis,
\newblock ``Wide residual networks,''
\newblock in {\em Proceedings of the British Machine Vision Conference (BMVC)},
  September 2016, pp. 87.1--87.12.

\end{thebibliography}
\end{small}

\end{document}